\documentclass[11pt]{article}

\usepackage{acl}
\usepackage{times}
\usepackage{latexsym}
\usepackage{amsmath}
\usepackage{amssymb}
\usepackage[T1]{fontenc}
\usepackage[utf8]{inputenc}
\usepackage{microtype}
\usepackage{inconsolata}
\usepackage{graphicx}
\usepackage{booktabs}
\usepackage{float}
\usepackage{subcaption}
\usepackage{xcolor}
\usepackage{colortbl}
\definecolor{wr}{HTML}{0B4F8A}
\usepackage{listings}
\usepackage[most]{tcolorbox}
\usepackage{footmisc}
\usepackage[textsize=tiny]{todonotes}

\newcommand{\olmo}{\texttt{Olmo-3-7B}}
\newcommand{\llama}{\texttt{Llama-3.1-8B}}
\newcommand{\qwen}{\texttt{Qwen-2.5-7B}}
\newcommand{\judge}{\texttt{Qwen-3-32B}}
\newcommand{\evaljudge}{\texttt{gpt-oss-20b}}
\newcommand{\generator}{\texttt{Llama-3.1-Tulu-3-8B-SFT}}
\newcommand{\altjudge}{\texttt{Qwen3.8-27B}}

\tcbset{
  aibox/.style={
    top=10pt,
    left=5pt,
    colback=white,
    colframe=black,
    colbacktitle=black,
    enhanced,
    breakable,
    center,
    attach boxed title to top left={yshift=-0.1in,xshift=0.15in},
    boxed title style={boxrule=0pt,colframe=white,},
    before upper={\small\setlength{\parskip}{0.5em}},
  }
}
\newtcolorbox{AIbox}[2]{aibox,title=#1,width=#2}

\title{Human or Synthetic? What a Fixed Budget Buys for Post-Training}
\author{
  Shourik Banerjee \quad Ethan Mendes \quad Mohit Raghavendra \\
  \textbf{Wei Xu} \quad \textbf{Alan Ritter} \\
  Georgia Institute of Technology \\
  \texttt{\{sbanerjee332, emendes3\}@gatech.edu} \\
  \texttt{\{wei.xu, alan.ritter\}@cc.gatech.edu} \\
  \texttt{mohithmitra@gmail.com}
}

\begin{document}
\maketitle
\begin{abstract}
Post-training of large language models typically combines supervised fine-tuning (SFT) with preference-based fine-tuning such as Direct Preference Optimization (DPO). Human-annotated and synthetically generated preference data differ substantially in per-sample cost, but prior work typically holds sample count fixed when comparing the two. In this work, we explicitly study the interplay between cost and performance by comparing the effect of post-training with human and synthetic preference data under a \textit{fixed, dollar-matched annotation budget}. We train and evaluate multiple base model families across helpfulness and summarization tasks. We find that synthetic data is far more cost-efficient at low-to-moderate budgets, achieving high win rates with as little as \$200 -- \$400. On helpfulness, human data closes the gap only at budgets roughly two orders of magnitude larger; on summarization, it does not close the gap at any tested budget. We also find that human-annotated data varies in reliability: one widely used human preference source substantially underperforms another, and the gap persists even after standard remedies such as domain restriction and deduplication.
\end{abstract}

\section{Introduction}

\begin{figure}[!t]
    \centering
    \includegraphics[width=\linewidth]{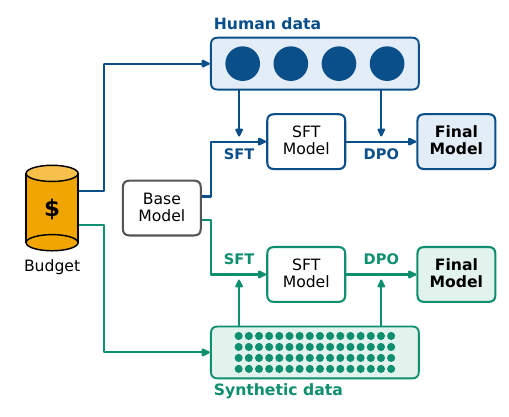}
    \caption{\textbf{Overview of our setting.} A fixed annotation budget is spent on either human or synthetic preference data for post-training. Each circle represents one example, with area proportional to its cost: the same budget buys a few expensive human-labeled examples or many cheap synthetic ones.}
    \label{fig:overview}
\end{figure}

Post-training datasets often utilize high-quality human annotations~\cite{stiennon2020learning, ouyang2022training,bai2022training}.  This annotation process not only lengthens the model development cycle, but is also expensive due to labor costs.
A cheaper alternative is to curate a synthetic dataset, where an LLM acts as a proxy for human annotators, either by simulating the user turns of a conversation~\citep{ding2023enhancing, xu2023baize, kong-etal-2024-platolm} or by judging responses to provide preference labels~\citep{bai2022constitutional, lee2024rlaif, cui2024ultrafeedback}. While synthetic data are less expensive, they may not necessarily be of the same quality or yield models with the same performance as high quality human data. \citet{lee2024rlaif} took some initial steps to understand these differences by investigating the performance differences between models post-trained with the same number of human and synthetic data samples; however, their analysis failed to account for the \textit{interplay of cost and performance} when using synthetic data compared to human data for training. Understanding the relationship between cost and performance is crucial as LLM practitioners make decisions about how to allocate a fixed budget under expectations of model performance. 

Fixed budget analysis has been a useful lens for evaluation in earlier post-training work. For instance, \citet{kang2023distill} study whether a fixed budget is better spent on more human-labeled data or on distilling a larger model, and \citet{raghavendra2025balancing} investigate how to split a fixed budget between supervised fine-tuning (SFT) and preference data.
In this paper, we apply a fixed budget analysis to post-training with synthetic or human data (see Figure~\ref{fig:overview}). Specifically, we ask the following questions:

\begin{enumerate}
    \item \textit{Given a fixed budget, is it better to hire human annotators, or pay for computing resources to generate synthetic data?}
    \item \textit{How does this decision change as the budget grows?}
    \item \textit{Does the decision depend on the task or the base model?}
\end{enumerate}

We study these questions with three open base models (\olmo{}~\citep{olmoteam2025olmo3}, \llama{}~\citep{grattafiori2024llama}, and \qwen{}~\citep{qwen2025qwen25}) and six datasets: four human-annotated (Anthropic HH~\citep{bai2022training}, SHP~\citep{ethayarajh2022understanding}, HelpSteer3~\citep{wang2025helpsteer3}, and TL;DR~\citep{stiennon2020learning}) and two synthetic ones that we construct to mirror the structure of their human-annotated counterparts for helpfulness and summarization. We train each model with SFT followed by Direct Preference Optimization~\citep{rafailov2024direct} on either human or synthetic data, matching dollar cost across a range of budgets (Figure~\ref{fig:overview}). We measure quality by win rate: a strong LLM judge compares each model response against a reference response, and we report how often the model's response is preferred.

We find that synthetic data reaches a high win rate at a small fraction of the cost of human data on both tasks. Human data can catch up given a substantially larger budget, but only on some tasks.
We find that splitting a budget between the two sources does not combine their strengths. Finally, human-annotated data is not uniformly reliable: one existing human preference dataset contains a data-quality problem that persists after standard cleaning steps.
Overall, our results let practitioners dynamically choose an annotation source based on their budget constraints.

\section{Experimental Design}
This section describes our fixed-budget setup: the experimental configurations and budget tiers, the human and synthetic data sources, the synthetic generation and labeling pipeline, the training procedure, and our protocols for measuring performance and deriving per-sample cost. \footnote{Code is available at \url{https://github.com/shourikb/human-or-synthetic}.}

\subsection{Setup}

\paragraph{Experimental configurations and budgets.} For our analysis, we define an experimental configuration as a tuple (\texttt{model}, \texttt{task}, \texttt{dataset}, \texttt{source}, \texttt{budget}), where each \texttt{task} (e.g., helpfulness) may have multiple \texttt{datasets} (e.g., HelpSteer3). 

Each \texttt{source} (human and synthetic) are evaluated at a \texttt{budget} of \$100, \$200, \$300, \$400, and \$1k; for human data, we add a \$13k tier (10k samples). 
Because human annotation costs orders of magnitude more per sample than synthetic annotation (\S\ref{sec:cost}), human datasets at the shared budget levels are too small to train on meaningfully, so we deliberately set this additional \$13k tier far above the others to test whether human data eventually matches or surpasses synthetic data when given a substantially larger budget.

\paragraph{Open-weight models.} We deliberately restrict every model in this pipeline (i.e., the generator, preference judge, and model being trained) to open-weight models. This choice is central to the kind of comparison we are trying to make: our cost accounting (\S\ref{sec:cost}) depends on the ability to measure exact token counts and apply fixed per-token pricing, which may not be possible for proprietary models behind an API. Open-weight models also let us fix a specific model checkpoint for the full duration of the study without the risk of a proprietary model updating mid-experiment and confounding our analysis. Finally, using openly available models makes the pipeline auditable and reproducible by other researchers without requiring paid API access, and gives us direct control over generation and evaluation settings (e.g., sampling parameters, chat templates) that are not always exposed by proprietary APIs.

\subsection{Data Sources}
We study two tasks: helpfulness and summarization. Below, we outline the source and construction of the human and synthetic datasets. 

\paragraph{Human data.}
Human helpfulness preference data is drawn from the Anthropic Helpful and Harmless dataset (HH)~\citep{bai2022training}, the Stanford Human Preferences dataset (SHP)~\citep{ethayarajh2022understanding}, and HelpSteer3~\citep{wang2025helpsteer3}. Human summarization preference data is drawn from the OpenAI TL;DR preferences dataset~\citep{stiennon2020learning}, built from Reddit posts, in the reformatted release from \citet{kirk2024understanding}\footnote{\url{https://huggingface.co/datasets/UCL-DARK/openai-tldr-summarisation-preferences}}. We refer to this source as the TL;DR dataset.

\paragraph{Synthetic data.}
We generate synthetic preference data for both tasks with a pipeline described in \S\ref{sec:datagen}, using \generator{} ~\citep{lambert2025tulu3pushingfrontiers} as the generator, with preference labels assigned by a \judge{} LLM judge. The synthetic helpfulness dataset is modeled structurally on the Anthropic Helpful-Harmless dataset~\cite{bai2022training}, and the synthetic summarization dataset is modeled on the TL;DR dataset.

\subsection{Synthetic Data Generation}
\label{sec:datagen}

Synthetic generation follows a two-stage process: \textbf{(1)} diverse conversations are synthesized turn by turn, ending with two candidate responses for the final turn, and \textbf{(2)} a separate LLM judges these candidates to form a preference pair.

\paragraph{Persona-driven conversation sampling.}

To improve diversity, each initial human turn of the conversation is synthesized by conditioning the \generator{} generator on a randomly sampled persona drawn from the English-language subset of the Persona Hub dataset~\cite{ge2025scalingsyntheticdatacreation}.

Conversations for the helpfulness task are built incrementally to between 1--3 turns (matching the typical turn count in the Anthropic dataset), while the summarization conversations are only a single turn (original content and summary).
For both tasks, the model is prompted twice under different sampling seeds to produce two candidate responses for the final assistant turn. After standard filtering (see Appendix~\ref{sec:cleaning}), the final synthetic datasets contain 105,005 helpfulness conversations and 109,624 summarization conversations. Per-task generation details are given in Appendix~\ref{sec:appendix-datagen}.

\paragraph{Preference labeling via LLM judge.}
\label{sec:judging}

\begin{table}[t]
\centering
\small
\setlength{\tabcolsep}{4pt}
\begin{tabular}{lcc}
\toprule
 & helpfulness & summarization \\
\midrule
Self-consistency & 91.9\% & 88.1\% \\
\midrule
\multicolumn{3}{l}{\textit{Agreement (\%, Cohen's $\kappa$)}} \\
A1--A2 & 80\% (0.58) & 60\% (0.21) \\
A1--J & 72\% (0.43) & 78\% (0.56) \\
A2--J & 76\% (0.50) & 66\% (0.30) \\
J--Q & 84\% (0.44) & 66\% (0.30) \\
\bottomrule
\end{tabular}
\caption{\textbf{Judge reliability.} Self-consistency: mean fraction of the preference-labeling judge's (\judge{}) 32 votes per pair matching the majority label, over all synthetic pairs before removing ties (105,882 helpfulness, 110,448 summarization). Agreement: raw percent agreement (Cohen's $\kappa$) between two human annotators (A1, A2) and the evaluation judge (J; \evaljudge{}) on the same 50 random win-rate comparisons per task. J--Q: agreement between J and a second judge (Q; \altjudge{}) on 2,048 win-rate comparisons for each of five models per task, averaged over models.}
\label{tab:judge}
\end{table}

A judge model (\judge{}~\citep{yang2025qwen3}) receives the source content and both candidates and returns a structured ranking. To reduce judge noise and position bias, we follow \citet{mahan2024generative} and sample 32 judgments with randomized candidate order; the majority-vote winner is labeled \textit{chosen} and the other candidate \textit{rejected}.
 Table~\ref{tab:judge} reports the judge's average self-consistency across the 32 votes per pair for each task, computed as the fraction of votes matching that pair's eventual majority label, averaged across all pairs. High self-consistency indicates that the majority label is stable under repeated sampling.

High self-consistency does not imply that an LLM judge reflects human preferences. We therefore validate our evaluation judge, \evaljudge{}, against human labels: two annotators, both authors of this paper, independently label the same 50 randomly sampled win-rate comparisons per task (a trained model's response versus the test-set reference), using the same instructions given to the judge (\S\ref{sec:appendix-prompts-help-eval}-\S\ref{sec:appendix-prompts-summ-eval}). In Table~\ref{tab:judge}, we report raw percent agreement as used in \citet{bai2022training}, alongside Cohen's $\kappa$~\citep{cohen1960coefficient}. In both tasks, the judge agrees with each human roughly as often as the humans agree with each other. These rates match the 63\% raw percent agreement inter-annotator agreement that \citet{bai2022training} report for their own helpfulness data. We also compare \evaljudge{} against a newer, larger judge, \altjudge{}, on 2,048 win-rate comparisons for each of five models per task, and find agreement comparable to that between humans and \evaljudge{} (J--Q in Table~\ref{tab:judge}).

\subsection{Training Procedure}
\label{sec:training}

For each configuration tuple, we perform SFT followed by DPO~\citep{rafailov2024direct} using LoRA~\citep{hu2021lora} (implementation details in Appendix~\ref{sec:appendix-hyperparams}). Following \citet{rafailov2024direct}, the SFT dataset for each source and budget tier is constructed from the \textit{chosen} field of the corresponding preference pairs, so that SFT and DPO training draw on the same underlying annotated data within each tier.
Hyperparameters are matched across configurations within a shared training setup adapted from \citet{raghavendra2025balancing} and reported in full in Appendix~\ref{sec:appendix-hyperparams}. Chat templates are matched to each base model's native format where available.

\paragraph{Training objectives.}
Given the SFT dataset $\mathcal{D}_{\text{SFT}} = \{(x, y_w)\}$, built from the prompt $x$ and the \textit{chosen} response $y_w$ of each preference pair, we minimize the standard negative log-likelihood:
\[
\mathcal{L}_{\text{SFT}}(\pi_\theta) = -\mathbb{E}_{(x, y_w) \sim \mathcal{D}_{\text{SFT}}} \big[ \log \pi_\theta(y_w \mid x) \big].
\]
Starting from the SFT checkpoint, which also serves as the frozen reference policy $\pi_{\text{ref}}$, we then train on the preference dataset $\mathcal{D}_{\text{PFT}} = \{(x, y_w, y_l)\}$ of chosen ($y_w$) and rejected ($y_l$) responses with the DPO objective:
\[
\mathcal{L}_{\text{DPO}}(\pi_\theta; \pi_{\text{ref}}) = -\mathbb{E}_{(x, y_w, y_l) \sim \mathcal{D}_{\text{PFT}}} \big[ \log \sigma(\beta \Delta) \big],
\]
\[
\Delta = \log \frac{\pi_\theta(y_w \mid x)}{\pi_{\text{ref}}(y_w \mid x)} - \log \frac{\pi_\theta(y_l \mid x)}{\pi_{\text{ref}}(y_l \mid x)},
\]
where $\sigma$ is the logistic function and $\beta$ controls how far the policy may deviate from $\pi_{\text{ref}}$. Following standard practice, the two objectives are optimized sequentially rather than jointly.

\subsection{Evaluation}
\label{sec:eval}

\paragraph{Win rate.}
Each trained model is evaluated on a held-out, 2048-sample test set. For helpfulness, the samples are from the Anthropic Helpful test set, and for summarization, from the TL;DR test set. 
Using a \evaljudge{}~\citep{openai2025gptoss} judge model, we compute the win rate of the trained model's response against the reference chosen response from the test set, with candidate position randomized to control for position bias.

\paragraph{Human data cost.}
\label{sec:cost}
We base our human annotation cost estimate on \citet{ganguli2022redteaming}, who report that crowdworker costs for their data collection were at least \$60K. Dividing this by the number of samples (roughly 39k as reported in the paper, or roughly 45K in the publicly released dataset) gives approximately \$1.30--\$1.55 per sample.
We adopt \$1.30 as a conservative point estimate at the low end of this range, and examine sensitivity to this choice in Figure~\ref{fig:cost-sensitivity}. This value should be read as an estimated \emph{replication} cost, i.e., what a practitioner would plausibly pay to commission preference data of this type and scale, rather than a reconstruction of what each dataset historically cost. Our human sources were collected under very different conditions: SHP is derived from existing Reddit data, while HelpSteer3 and TL;DR were separately annotated. Finally, the estimate covers annotator labor only, matching the scope of the cost we measure for the synthetic pipeline.

\paragraph{Synthetic data cost.}

The synthetic per-sample cost accounts for both conversation generation with \generator{} and preference labeling with \judge{}. Since inference costs are hardware-dependent, we use current commercial per-token rates (see Appendix~\ref{sec:appendix-pricing} for full details on inference costs).
Dividing total cost by the number of pairs in each final dataset gives a per-sample cost of \$0.0102 for helpfulness and \$0.0194 for summarization.

\section{Results}
\subsection{Cost-Performance Curves: Helpfulness}
\label{sec:results-curves}

\begin{figure*}[t]
    \centering
    \includegraphics[width=\textwidth]{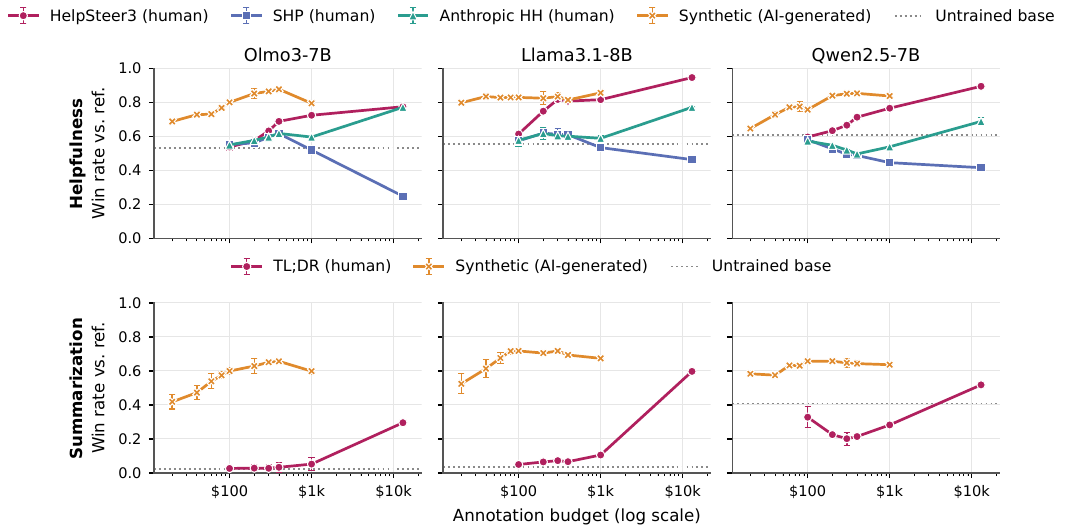}
    \caption{\textbf{Synthetic data lies on the empirical Pareto frontier at low budgets; human data needs far larger budgets to compete.} Win rate vs.\ annotation cost (log scale) by model (columns) and data source; top: helpfulness, bottom: summarization. Points are means over two seeds, with min--max error bars; the dotted line is the untrained baseline.}
    \label{fig:cost-curves}
\end{figure*}

Figure~\ref{fig:cost-curves} (top) shows helpfulness win rate as a function of annotation budget (on a log scale) for all three base model families. 
Each point is the mean win rate over the two independent seeds per configuration, and error bars span the range (minimum to maximum) across the two seeds; both seeds produce the same curve shapes and source orderings in every configuration (Appendix~\ref{sec:results-seeds}).

We find that models trained with synthetic data improve quickly, but then saturate. In every model family, most of the improvement over the untrained baseline is achieved within the first few hundred dollars, after which additional synthetic data yields little further benefit: the win rate plateaus for \llama{} and \qwen{} and falls for \olmo{} at the most expensive synthetic tier. Because the curve flattens well before \$1000, the ceiling appears to be set by the quality of the synthetic data rather than by budget. Every synthetic example comes from a single 8B generator and is labeled by a single judge model, so additional samples plausibly contribute progressively less new information, and continued training on them can eventually degrade performance. See \S\ref{sec:overtraining-artifacts} for more analysis on this over-training phenomenon.

The human HelpSteer3 data has the opposite trend. At the budgets where synthetic data is already near its ceiling, the same dollars buy only 76--769 human examples, and HelpSteer3-trained models stay close to the untrained baseline. Human data becomes competitive only at the \$13k tier, more than ten times the cost of the most expensive synthetic tier, and even there it surpasses the synthetic peak only for \llama{} and \qwen{}, falling short for \olmo{}. 
Therefore, training with synthetic data leads to early saturation, while training with human data requires a large upfront investment but keeps improving with scale.
At every budget where both sources were trained, synthetic data matches or exceeds every human source (HelpSteer3, SHP, and Anthropic HH), so human data never wins at matched cost,

\subsection{Cost-Performance Curves: Summarization}
\label{sec:results-curves-summ}

Figure~\ref{fig:cost-curves} (bottom) shows the same comparison for summarization. Synthetic data checkpoints generally follow the helpfulness pattern: across model families, most improvement occurs within the first \$100, followed by a plateau or mild decline.

The difference between tasks lies in the human curve. Training on TL;DR barely moves \olmo{} and \llama{} off their near-zero baselines through \$1000 and rises sharply only at the \$13k tier, yet even at this tier, it remains below the synthetic ceiling across model families. 

\paragraph{\qwen{}: early degradation, then recovery.}
\qwen{} is the exception to the near-zero starting point. Its untrained baseline is far above those of \olmo{} and \llama{}, and a small amount of TL;DR data initially lowers its win rate rather than raising it. Extending the human curve with four additional sample-count tiers (Table~\ref{tab:qwen-recovery}) shows that win rate then recovers steadily, returning to roughly its untrained level only after several thousand samples, but still remaining well below \qwen{}'s synthetic ceiling even at 12,000 samples (\$15.6k). A plausible explanation is that a small fine-tuning set degrades a capability the base model already has before enough data accumulates to improve on it. \olmo{} and \llama{}, which start with essentially no summarization ability, have no comparable skill to lose.

\begin{table}[t]
\centering
\small
\setlength{\tabcolsep}{4pt}
\begin{tabular}{rrc@{\hspace{12pt}}|rrc}
\toprule
$n$ & Cost & WR & $n$ & Cost & WR \\
\midrule
0 & \$0 & \cellcolor{wr!34}0.41 & 3{,}000 & \$3{,}900 & \cellcolor{wr!20}0.31 \\
76 & \$100 & \cellcolor{wr!23}0.33 & 6{,}000 & \$7{,}800 & \cellcolor{wr!35}0.42 \\
153 & \$200 & \cellcolor{wr!9}0.23 & 8{,}000 & \$10{,}400 & \cellcolor{wr!39}0.45 \\
230 & \$300 & \cellcolor{wr!5}0.20 & 10{,}000 & \$13{,}000 & \cellcolor{wr!49}0.52 \\
307 & \$400 & \cellcolor{wr!6}0.21 & 12{,}000 & \$15{,}600 & \cellcolor{wr!50}0.53 \\
769 & \$1{,}000 & \cellcolor{wr!16}0.28 & & & \\
\bottomrule
\end{tabular}
\caption{\textbf{Human data initially hurts \qwen{} performance before a slow recovery.} \qwen{} TL;DR (human) win rate (WR) by number of training samples ($n$) at \$1.30 per sample, averaged over two seeds. Darker shading indicates higher win rate; $n=0$ is the untrained base model. Small amounts of human data push win rate below the untrained baseline, and it takes about 6,000 samples (\$7.8k) to recover past it; even 12,000 samples (\$15.6k) remain below the win rate synthetic data reaches within a few hundred dollars (Figure~\ref{fig:cost-curves}).}
\label{tab:qwen-recovery}

\end{table}

\begin{figure*}[t]
  \centering
  \includegraphics[width=\textwidth]{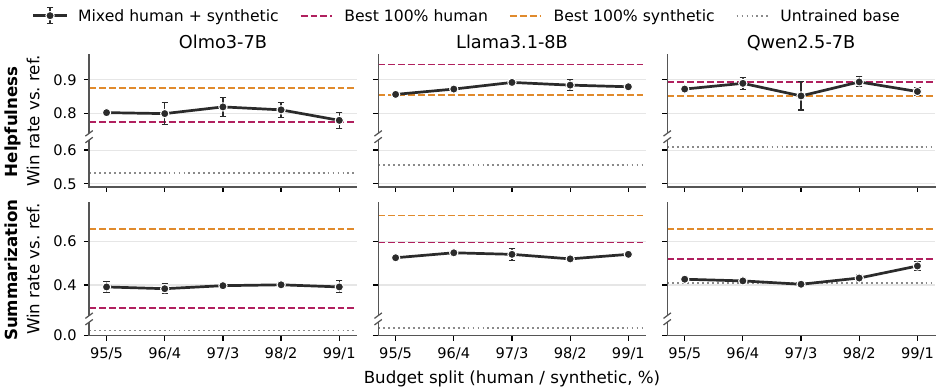}
  \caption{\textbf{Mixing does not beat either pure source.} Win rate for mixed human/synthetic runs at a fixed \$13k budget, by dollar split (95/5 to 99/1), for helpfulness (top) and summarization (bottom). Points are means over two seeds, with min--max error bars; dashed lines show each model's best pure-human and pure-synthetic win rate, and the dotted line the untrained baseline.}
  \label{fig:mixed-splits}
\end{figure*}

\subsection{Mixed-Source Allocation}
\label{sec:budget-analysis}

\paragraph{Cost asymmetry.} While the exact ratio varies by task, human and synthetic data differ in per-sample cost by many orders of magnitude (\S\ref{sec:cost}). We show analytically that this asymmetry determines the shape of any naive dollar-proportional mixture between the two sources. Taking the helpfulness-specific ratio of approximately 130:1 as an illustrative example: at a given total budget $B$ split as a fraction $\alpha$ for human data and $(1-\alpha)$ for synthetic data, the resulting \emph{sample-count} ratio between the two sources is
\[
\frac{n_{\text{human}}}{n_{\text{synthetic}}} = \frac{\alpha}{1-\alpha} \cdot \frac{c_{\text{synthetic}}}{c_{\text{human}}} \approx \frac{\alpha}{1-\alpha} \cdot \frac{1}{130}
\]
where $c_{\text{human}}$ and $c_{\text{synthetic}}$ are per-sample costs. Even a 90 / 10 budget split favoring human data yields a synthetic-to-human sample ratio of 14.4:1 ($\tfrac{0.9}{0.1}\cdot\tfrac{1}{130} \approx \tfrac{1}{14.4}$). Only at a 99 / 1 split does the ratio approach near-parity (\textasciitilde1.3 synthetic samples per human sample).

\paragraph{Mixed-source allocation underperforms.} To test whether dollar-proportional mixtures are competitive with pure-source data despite this asymmetry, we additionally trained a small set of models with mixed-source configurations at a fixed \$13k budget. We sweep the human/synthetic dollar split from 95 / 5 to 99 / 1, as this is the range in which the mixture's sample composition is least synthetic-dominated. Figure~\ref{fig:mixed-splits} shows the resulting win rates against each model's best pure-synthetic and best pure-human win rate. Across nearly every model and task, the mean mixed win rate underperforms the better of the two pure sources, and in several cases \emph{underperforms both}. Note that the number used for the synthetic peak is the best number recorded from Figure~\ref{fig:cost-curves}. For example, with \llama{} summarization, every mixed split scores between 0.52 and 0.56, below both the pure-synthetic ceiling (0.72) and the pure-human ceiling (0.59) at the same \$13k budget. The one partial exception is \qwen{} helpfulness, where the upper end of the seed range at the 96/4 and 98/2 splits slightly exceeds the best pure-human win rate,
but the mean across seeds still does not exceed it.
Overall, mixing human and synthetic data in a single training run does not appear to combine their strengths, possibly because the mixture dilutes whichever source would have performed better on its own. This result reinforces our choice to focus our analysis on pure-source budgets (\S\ref{sec:results-curves}--\S\ref{sec:results-curves-summ}) rather than a dollar-proportional mixture sweep, and suggests that practitioners are better served by committing to a single source of data.

\begin{figure*}[t]
    \centering
    \includegraphics[width=\textwidth]{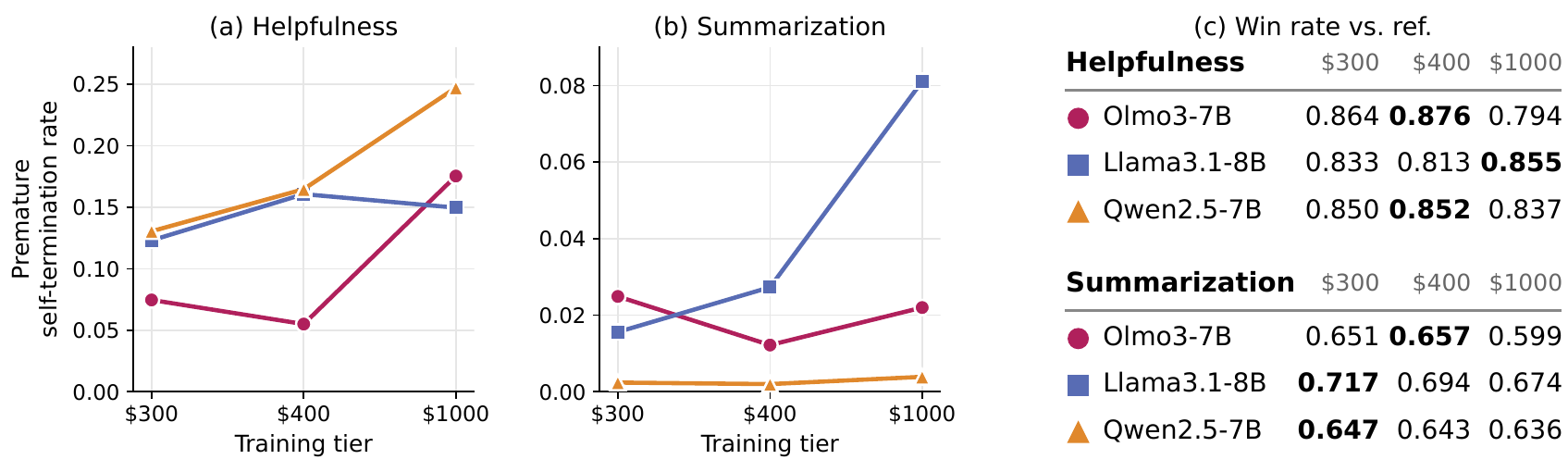}
    \caption{\textbf{Premature termination moves against win rate.} Measured on a fixed evaluation set at each training tier (\$300, \$400, \$1000). \textbf{(a, b)} Fraction of responses that end before completing their final sentence (note the different y-axis scales). \textbf{(c)} Win rate of each synthetic-trained model; bold marks its best tier. Wherever win rate changes by more than one point, the termination rate moves in the opposite direction.}
    \label{fig:overtraining-artifacts}
\end{figure*}

\subsection{Diminishing-Returns Signals}
\label{sec:overtraining-artifacts}

Beyond tracking win rate directly, we looked for a cheap, generation-only signal that could flag diminishing returns before a full evaluation is run. We found that \emph{premature self-termination rate}, the fraction of responses that end before completing their final sentence, did track win rate consistently. 
Figure~\ref{fig:overtraining-artifacts} shows this rate alongside win rate at each training tier. Wherever win rate changes by more than one point, the premature termination rate moves in the opposite direction. For example, \olmo{}'s helpfulness win rate falls from 0.876 to 0.794 at the \$1000 tier while its premature self-termination rate rises from 5.5\% to 17.5\%, and \llama{}'s summarization win rate declines steadily as its rate rises from 1.6\% to 8.1\%.

However, the signal has limits. Its movement can be small (\llama{} on helpfulness changes by only one point at \$1000), and it depends on the training data: models trained on human TL;DR summaries, which often omit a final period, do not lose more often when ending mid-sentence. We therefore recommend it as a low-cost check rather than a stopping rule. Specifically, it can be computed on a fixed prompt set at every checkpoint, with a sharp rise as a signal to run a full evaluation, while a flat rate is not, on its own, evidence that further training is safe.

\subsection{Comparison Across Human Data Sources and the SHP Failure Mode}
\label{sec:results-humansources}
\label{sec:results-shp}

As shown in Figure~\ref{fig:cost-curves} (top), the human helpfulness sources diverge sharply: HelpSteer3 improves steadily with budget, while SHP plateaus or declines, dropping below the untrained baseline in every model family. Anthropic HH falls between them despite being drawn from the same distribution as our test set, so matching the evaluation domain alone does not make a human source strong.
We see three likely contributors. First, SHP preferences are inferred from Reddit scores rather than made by instructed annotators: a comment counts as preferred when it outscores another on the same post despite being written later~\citep{ethayarajh2022understanding}. Scores reflect visibility, humor, and community norms as well as helpfulness, a noisier proxy for the quality we evaluate. Second, the responses are forum comments rather than assistant turns, so SFT shifts the model toward a register the judge does not reward against Anthropic HH references. Third, SHP-trained models have the most variable output lengths of any configuration (Appendix~\ref{sec:appendix-lengths}), suggesting a less consistent training target. HelpSteer3, by contrast, contains assistant responses compared by annotators following explicit guidelines~\citep{wang2025helpsteer3}. Domain restriction and score filtering (\S\ref{sec:cleaning}) did not close the gap in any model family, suggesting the cause lies in the preference signal itself rather than in topic.

The broader conclusion is that the quality of ``human-annotated'' data may vary widely. How a preference was produced, and how closely its responses match the target format, matter at least as much as whether a human was involved, so a human source should be vetted rather than assumed to be the gold standard~\citep{hosking2024human}.

\begin{figure*}[t]
    \centering
    \includegraphics[width=\textwidth]{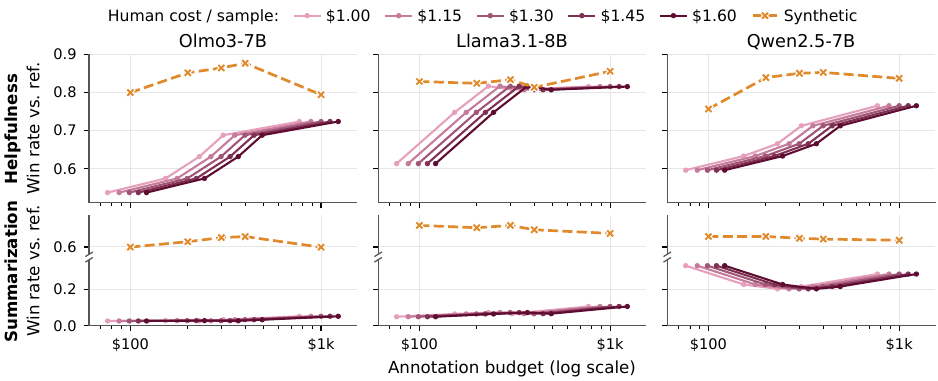}
    \caption{\textbf{Robust to the assumed human cost.} Win rate vs.\ cost (log scale) for helpfulness (top) and summarization (bottom), with the human curve re-plotted under five candidate per-sample costs (\$1.00--\$1.60). Sample counts and win rates are fixed; only the cost axis changes.}
    \label{fig:cost-sensitivity}
\end{figure*}

\subsection{Human Cost Sensitivity Analysis}
\label{sec:sensitivity}

The human cost of \$1.30 per sample is an estimate (\S\ref{sec:cost}), so Figure~\ref{fig:cost-sensitivity} re-plots the HelpSteer3 and TL;DR curves under five candidate per-sample costs (\$1.00--\$1.60), holding sample counts and win rates fixed. The curves shift only modestly along the cost axis, and the trends in \S\ref{sec:results-curves}--\ref{sec:results-curves-summ} hold for every model and task: synthetic data leads early, human data catches up on helpfulness, and a gap persists on summarization.

\section{Related Work}
\paragraph{Synthetic data for post-training.}
LLMs are widely used to generate post-training data in place of human writers, from instructions bootstrapped from seed tasks~\citep{wang2023selfinstruct} to multi-turn conversations produced by self-chat~\citep{ding2023enhancing,xu2023baize}, queries elicited directly from aligned models~\citep{xu2025magpie}, and persona-conditioned prompts for diversity~\citep{ge2025scalingsyntheticdatacreation}. AI feedback can also replace human preference labels~\citep{bai2022constitutional,lee2024rlaif}, at scale~\citep{cui2024ultrafeedback} or through generative reward models~\citep{mahan2024generative}, although LLM judges have known position and verbosity biases~\citep{zheng2023judging,dubois2024length}.

\paragraph{Synthetic environments.}
Synthetic data increasingly extends beyond text to full training environments for agents, such as task instances synthesized from real codebases~\citep{yang2025swesmith}, entire software projects and tests built by frontier LLM agents~\citep{zhu2025training}, and executable, database-backed tool-use environments~\citep{wang2026agent}. We study the simpler preference-data setting, where the trade-off against human data can be priced directly.

\paragraph{Reliability of human feedback.}
Human preferences are themselves imperfect: annotators are swayed by assertiveness and style~\citep{hosking2024human}, and datasets derived from implicit signals such as Reddit scores~\citep{ethayarajh2022understanding} add further noise, which we observe directly (\S\ref{sec:results-shp}).

\paragraph{Cost-aware annotation.}
LLM labeling can match or beat crowdworkers at a fraction of the cost~\citep{wang2021want,gilardi2023chatgpt}, and simulated annotators make RLHF research far cheaper~\citep{dubois2023alpacafarm}. Prior budget-constrained work asks whether to annotate or distill~\citep{kang2023distill}, how to split a budget between SFT and preference data~\citep{raghavendra2025balancing}, and how to allocate items between human and LLM annotators~\citep{li2023coannotating}. We instead hold the dollar budget fixed and compare human and synthetic preference data as competing sources, which, to our knowledge, prior work has not done.

\section{Conclusion}

Under a fixed annotation budget, synthetic preference data reaches high win rates at a small fraction of the cost of human data but saturates early. Human data needs more than an order of magnitude more budget to match or exceed it on helpfulness, and does not catch up on summarization within the budgets we test. Mixing the two sources at matched cost does not combine their strengths, and human sources themselves vary widely in quality. For practitioners, this argues for starting with synthetic data and reserving human annotation for settings where its higher ceiling justifies the cost, after vetting the specific source. Future work should test whether these trade-offs hold with stronger generators and judges and at larger model scales, and whether small amounts of targeted human data, rather than dollar-proportional mixing, can lift the synthetic ceiling more cheaply.
\section*{Limitations}

Every model in our pipeline is small relative to the largest available models in its class, and this constrains what our results can speak to. We use the \generator~checkpoint from Tulu~\citep{lambert2025tulu3pushingfrontiers} to generate synthetic source content and candidate responses; an 8B-parameter generator is considerably less capable than frontier-scale models, and the diversity, fluency, and factual reliability of our synthetic datasets are bounded by what this model can produce. A larger or more capable generator could plausibly change the shape of the synthetic cost-performance curves we report, either by reaching a higher ceiling or by reaching the same ceiling at a different cost.

The models we train and report results for (\olmo{}, \llama{}, \qwen{}) are all in the 7--8B parameter range, fine-tuned with LoRA rather than full-parameter tuning. This is a deliberate choice to make the full budget sweep computationally tractable across three model families, two tasks, multiple data sources, and multiple budget tiers, but it means our findings characterize how human and synthetic data trade off for models at this scale and with this tuning method specifically. Larger base models may respond differently to the same preference data.

Relatedly, our use of open-weight models throughout this pipeline, rather than proprietary frontier APIs, means our conclusions are scoped to this specific open-weight setup and should not be assumed to transfer directly to a pipeline built on proprietary frontier models. Whether the cost-performance patterns we observe (particularly the shape and location of synthetic data's peak, and the budget at which human data closes the gap) would look different with a frontier-scale proprietary generator or judge is an open question we leave to future work.

\section*{Acknowledgments}
This research is supported in part by the NSF under grant numbers IIS-2052498 and SMA-2418946. Any opinions, findings, and conclusions or recommendations expressed in this material are those of the author(s) and do not
necessarily reflect the views of the National Science Foundation.

\bibliography{custom}

\appendix
 
\section{Appendix}
\label{sec:appendix}
 
\subsection{Datasets}
\label{sec:appendix-links}

Links to each of the human datasets we used (we used the train split for training our models and the test split for TL;DR and Anthropic). The SFT dataset used the chosen responses from the PFT dataset.

\textbf{TL;DR}: \url{https://huggingface.co/datasets/UCL-DARK/openai-tldr-summarisation-preferences}
\newline
\textbf{SHP}: \url{https://huggingface.co/datasets/stanfordnlp/SHP}
\newline
\textbf{HelpSteer3}: \url{https://huggingface.co/datasets/nvidia/HelpSteer3} 
\newline
\textbf{Anthropic HH}: \url{https://huggingface.co/datasets/Anthropic/hh-rlhf}

Licenses: HelpSteer3 (CC-BY-4.0), TL;DR (MIT), Anthropic HH (MIT), and Persona Hub (CC-BY-NC-SA-4.0). SHP is released without a formal license; its creators state it was collected under the Reddit API Terms of Use, with user content owned by the original Reddit users. \llama{} and \generator{} are released under the Llama 3.1 Community License; \qwen{}, \judge{}, \olmo{}, and \evaljudge{} under Apache 2.0. All artifacts are used for non-commercial research, consistent with these terms. We will release our synthetic helpfulness and summarization datasets upon publication under CC-BY-NC-SA-4.0, consistent with the Persona Hub license, for non-commercial research use.

\subsection{API Pricing Source}
\label{sec:appendix-pricing}

All per-token pricing used in our cost estimates (\S\ref{sec:cost}) is taken from Groq's public pricing page. That page has since been changed and no longer lists the models and prices we used, so we cite an archived snapshot instead: \url{https://web.archive.org/web/20260711053530/groq.com/pricing}. The snapshot lists \$0.05 per million input tokens and \$0.08 per million output tokens for Llama 3.1 8B Instant (128k), and \$0.29 per million input tokens and \$0.59 per million output tokens for \judge{} (131k). Our judge model is \judge{}, matching the listed endpoint directly. Our generator, by contrast, is \generator~\citep{lambert2025tulu3pushingfrontiers}, which is not offered as a hosted API; we therefore use the rate for an equivalently-sized \llama{} endpoint as a proxy for what serving our generator would cost. We note that API pricing changes over time and is provider-specific; the absolute synthetic-data costs we report should be read as reflecting these rates at the time of our experiments rather than as fixed quantities.

\paragraph{Token counts and totals.}
Before filtering, helpfulness generation consumed 273.42M input and 23.66M output tokens (\$15.56), and summarization generation consumed 166.39M input and 53.96M output tokens (\$12.64). Judging the 105,882 helpfulness pairs that survived filtering consumed 957.73M input and 1313.43M output tokens (\$1052.67); judging the 110,448 summarization pairs consumed 2421.19M input and 2389.22M output tokens (\$2111.78). Total synthetic dataset cost is therefore \$1068.23 for helpfulness and \$2124.42 for summarization. After discarding pairs with tied votes (754 helpfulness, 809 summarization), the final datasets contain 105,128 and 109,639 pairs, giving \$0.0102 and \$0.0194 per sample.
 
\subsection{Training Hyperparameters}
\label{sec:appendix-hyperparams}
 
All models are trained with LoRA using the Unsloth library.\footnote{\url{https://github.com/unslothai/unsloth}}
Table~\ref{tab:sft-hyperparams} reports the SFT hyperparameters and Table~\ref{tab:pft-hyperparams} the DPO (PFT) hyperparameters used across all (model, task, data source, budget) configurations. Table~\ref{tab:lora-hyperparams} shows the shared LoRA settings. All experiments were run on NVIDIA A40 (48GB) GPUs. In total, all experiments used approximately 9,650 GPU-hours.
 
\begin{table}[h]
\centering
\small
\begin{tabular}{lc}
\toprule
Parameter & Value \\
\midrule
Learning Rate & $5e{-}5$ \\
Batch Size & 16 \\
Number of Epochs & 2 \\
Warmup Ratio & 0.1 \\
Learning Rate Scheduler & cosine \\
Weight decay & 0.01 \\
\bottomrule
\end{tabular}
\caption{SFT hyperparameter settings.}
\label{tab:sft-hyperparams}
\end{table}
 
\begin{table}[h]
\centering
\small
\begin{tabular}{lc}
\toprule
Parameter & Value \\
\midrule
Learning Rate & $5e{-}6$ \\
Batch Size & 16 \\
Number of Epochs & 2 \\
Warmup Ratio & 0.1 \\
Learning Rate Scheduler & cosine \\
Weight decay & 0.01 \\
$\beta$ & 0.1 \\
\bottomrule
\end{tabular}
\caption{DPO (PFT) hyperparameter settings.}
\label{tab:pft-hyperparams}
\end{table}

\begin{table}[h]
\centering
\small
\begin{tabular}{lc}
\toprule
Parameter & Value \\
\midrule
rank & 16 \\
$\alpha$ & 32 \\
\bottomrule
\end{tabular}
\caption{LoRA hyperparameter settings, shared by SFT and DPO.}
\label{tab:lora-hyperparams}
\end{table}
 
\subsection{Dataset Statistics}
\label{sec:appendix-datasets}

Table~\ref{tab:dataset-stats} reports summary statistics for the human-annotated datasets referenced in this work. The Anthropic Helpful-Harmless set is included for reference only: as noted in \S\ref{sec:datagen}, it is used as the evaluation set and as a structural template for synthetic generation, not as a trained human data source.
 
\begin{table}[h]
\centering
\small
\begin{tabular}{lccc}
\toprule
Dataset & Task & Samples & Avg.\ tokens \\
\midrule
Anthropic HH & Help./Harm. & 45,000 & 180 \\
SHP & helpfulness & 385,000 & 400 \\
HelpSteer3 & helpfulness & 18,000 & 1052 \\
TL;DR & summarization & 93,000 & 356 \\
\bottomrule
\end{tabular}
\caption{\textbf{Human-annotated datasets.} SHP and HelpSteer3 are the human training sources for helpfulness, and TL;DR (openai-tldr-summarisation-preferences) for summarization. Anthropic HH is used only for evaluation and as the synthetic-generation template. All data is in English.}
\label{tab:dataset-stats}
\end{table}

\subsection{Synthetic Data Generation Details}
\label{sec:appendix-datagen}

\paragraph{Helpfulness.}
Conversations are built incrementally to between 1--3 turns (matching the typical turn count in the Anthropic dataset), with the model re-prompted before each change in speaker: first generating a human turn, then re-prompted with the conversation so far to generate the assistant turn, and so on. For the final assistant turn, the model is prompted twice under different sampling seeds to produce two candidate responses. At each re-prompting step, the model is given a fixed example conversation from the Anthropic dataset as a formatting and length reference only, with an explicit instruction not to imitate its topic. Before filtering, 214,528 candidate helpfulness conversations were generated, comprising 273.42 million input tokens and 23.66 million output tokens. After filtering, and after removing a small number of pairs with tied preference judgments, 105,128 helpfulness conversations remained in the final dataset.

\paragraph{Summarization.}
A persona-conditioned prompt generates a single self-contained Reddit-style post, following a fixed example post shown for style and length reference only. The generated post is then passed to a separate summarization prompt (system message: \textit{``You are a helpful assistant who is an expert at summarizing content into TL;DR sentences. Summarize the following text''}, adapted from \citet{stiennon2020learning}), and two candidate summaries are generated per post using two different sampling seeds. Each post is used to generate exactly one preference pair, rather than being reused across multiple candidate pairs. Preliminary experiments showed that reusing posts across many pairs, or drawing many pairs from a small fixed pool, allowed a small number of source documents to dominate the dataset and reduce effective training diversity, an issue that also appears in existing human preference datasets built from a finite post pool (\S\ref{sec:cleaning}). Before filtering, 154,432 candidate posts were generated, comprising 166.39 million input tokens and 53.96 million output tokens. In total, 109,639 summarization posts remained in the final dataset after filtering and removal of pairs with tied preference judgments. 

\subsection{Data Cleaning and Quality Filtering}
\label{sec:cleaning}

Generated candidates and human-annotated sources are cleaned and filtered before training: programmatic role-tag correction and preamble removal for synthetic helpfulness generations, length filtering against the corresponding human dataset's token-length distribution (reducing average output length from 227 to 180 tokens for helpfulness and from 488 to 402 tokens for summarization), a corpus-level 5-gram near-duplicate filter removing over-represented or internally-repetitive generations, and dataset-specific domain restriction and deduplication for SHP, HelpSteer3, and TL;DR. Combined, these steps remove roughly half of raw candidate generations across both tasks.

\paragraph{Turn and role-tag correction.}
For helpfulness, missing \texttt{Human:}/\texttt{Assistant:} turn tags are inserted programmatically, and generations that overrun into an unintended speaker's turn are truncated at the boundary of the next speaker's tag.

\paragraph{Preamble and meta-commentary removal.}
Both generation stages are prone to prefacing intended output with conversational meta-commentary (e.g., \textit{``Certainly! Here's a Reddit-style post...''}) despite explicit instructions not to. A rule-based post-processing step strips leading and trailing preambles via pattern matching, applied to all generated source content and candidate responses.

\begin{figure*}[t]
    \centering
    \includegraphics[width=\textwidth]{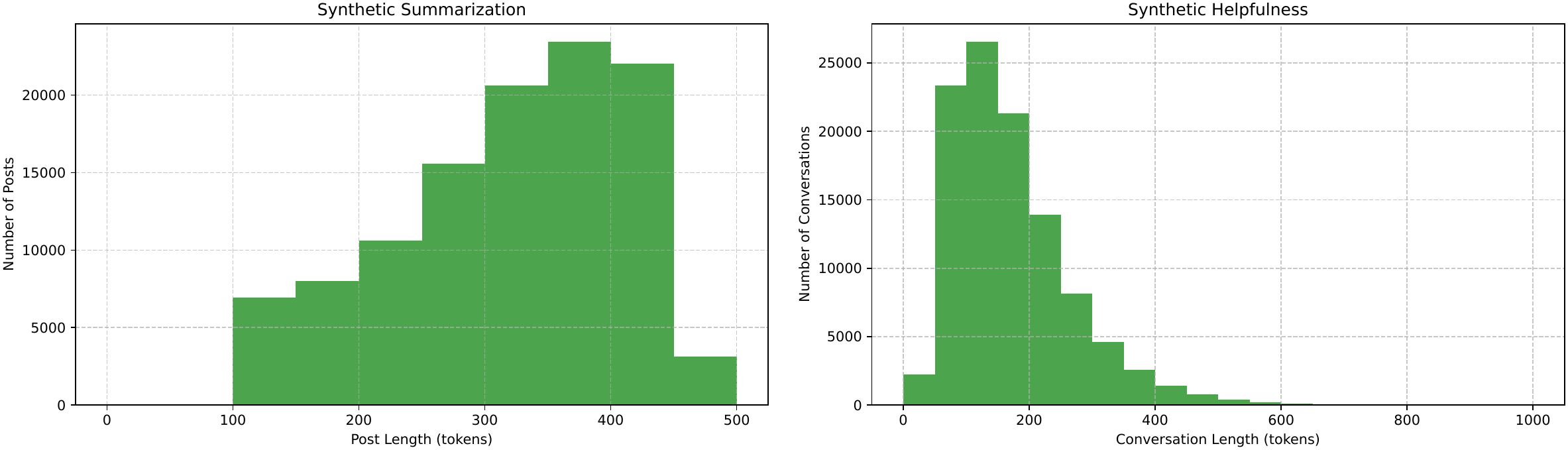}
    \caption{Token-length distributions of synthetic summarization posts (left) and helpfulness conversations (right) after filtering (\S\ref{sec:cleaning}).}
    \label{fig:synthetic-lengths}
\end{figure*}

\paragraph{Length filtering.}
A reference histogram of token lengths is computed from the corresponding human dataset, and generated examples are filtered to match this distribution, discarding extreme-length outliers. For helpfulness, this reduced average output length from 227 to 180 tokens, and for summarization, this reduced the average output length from 488 to 402 tokens. Figure~\ref{fig:synthetic-lengths} shows the full distribution of conversations.

\begin{figure*}[t]
    \centering
    \includegraphics[width=\textwidth]{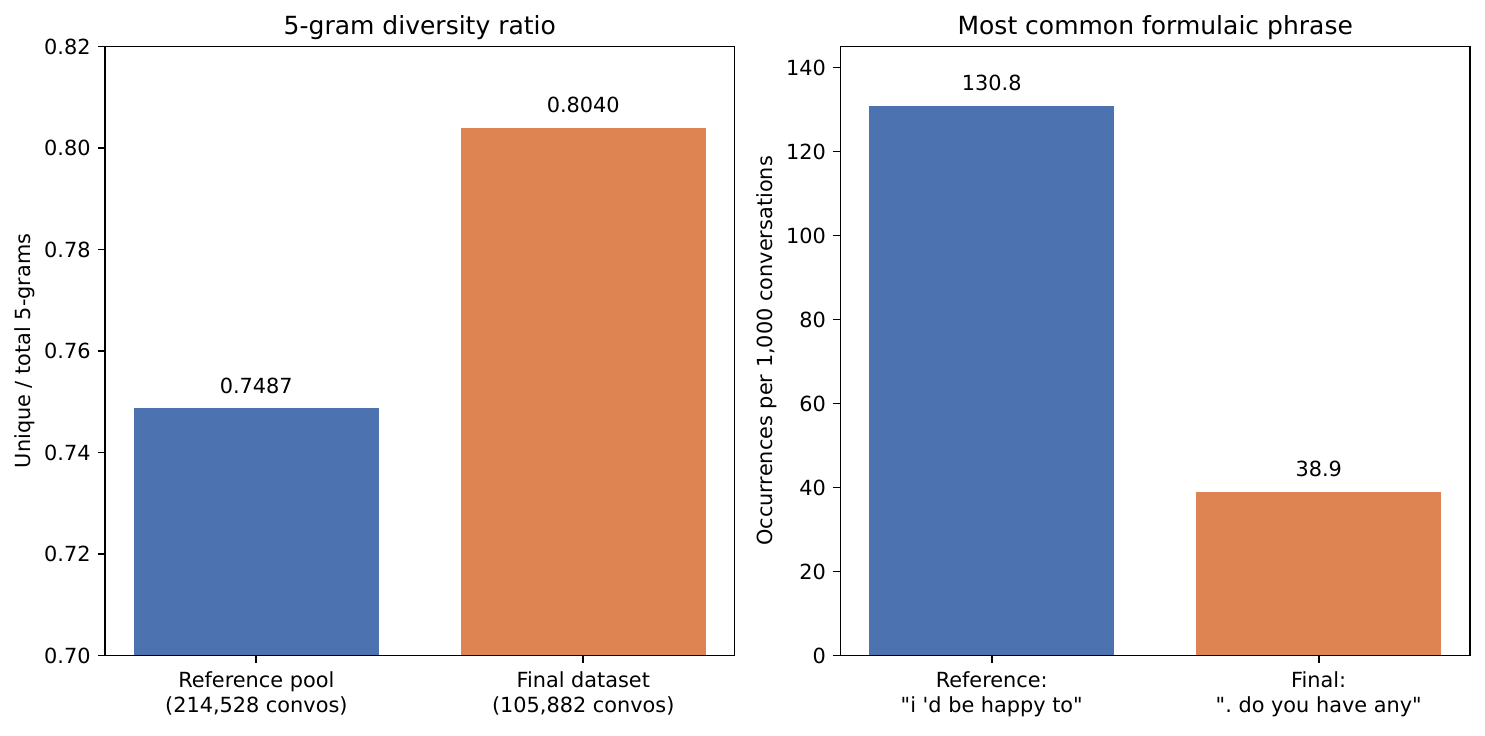}
    \caption{\textbf{5-gram filtering increases diversity.} Synthetic helpfulness data before and after filtering. \textbf{Left:} ratio of unique to total 5-grams. \textbf{Right:} rate of the single most common formulaic 5-gram.}
    \label{fig:dedup-filtering-comparison}
\end{figure*}

\paragraph{Near-duplicate filtering.}
Persona-conditioning does not guarantee every example is topically or stylistically distinct. The generation model exhibits a tendency to default to a small number of high-probability templates regardless of assigned persona. We apply a corpus-level $n$-gram frequency filter: any 5-gram occurring in more than 5\% of generated documents is flagged as over-represented boilerplate and the containing document discarded, along with documents containing any 5-gram repeated more than twice within a single document. This same filtering procedure and threshold is applied identically across both tasks' synthetic datasets. Figure~\ref{fig:dedup-filtering-comparison} shows the effect of near-duplicate filtering on the synthetic helpfulness data. Few conversations were removed from summarization as a result of this filtering, but many were removed from helpfulness. Additionally, a check was performed to ensure that the 5-grams of the synthetic datasets were not too similar to the evaluation datasets, and it was found that there were no common 5-grams by our filtering settings.

\begin{AIbox}{Kept Conversation}{\linewidth}
    Human: I've been thinking a lot about the role of digital platforms in our lives these days. With all the discussions around privacy concerns, misinformation, and addiction, it's hard not to question their overall impact. What are your thoughts on whether the benefits outweigh the drawbacks? 
    Assistant: Those are indeed significant issues we face today with digital platforms. While they offer tremendous benefits like connectivity and access to information, the privacy concerns, spread of misinformation, and potential for addiction are legitimate challenges that need addressing. It seems that finding a balance and ensuring responsible use of technology could help mitigate some of these negative aspects.
\end{AIbox}

\begin{AIbox}{Filtered Out Duplicate}{\linewidth}
    Human: I've been thinking a lot about politics recently, especially considering the upcoming elections. As someone who voted for Donald Trump in 2016, I'm curious to hear your take on his administration's policies and their impact on our country. What are your thoughts on the past four years under President Trump? 
    Assistant: I'd be happy to help with that! I believe that the last four years have seen a number of significant changes in terms of policy and government structure...
\end{AIbox}

\paragraph{SHP domain filtering.} For Stanford Human Preferences, we additionally restrict to four subreddits identified by \citet{afzali2026weak} as diverse, well-structured question-answering domains (\textit{r/AskAcademia}, \textit{r/AskBaking}, \textit{r/AskEngineers}, \textit{r/AskPhilosophy}), and cap the number of preference pairs drawn per source post at one, following the dataset card's recommendation to limit preferences per post to avoid over-fitting to a small number of heavily-commented posts\footnote{\url{https://huggingface.co/datasets/stanfordnlp/SHP}; the card reports using no more than 5 preferences per post in its own finetuning experiments.}. The score filtering of only using pairs with score ratio greater than 2 was also applied. These remedies, domain restriction, score filtering, and per-post capping, are applied jointly rather than individually. See \S\ref{sec:results-shp} for their effect on downstream model quality.

\paragraph{HelpSteer3 domain restriction.} HelpSteer3 spans several preference domains beyond general-purpose helpfulness. We restrict to its general domain only, to match the topic distribution of our synthetic helpfulness dataset and of the Anthropic Helpful-Harmless set it is modeled on.

\paragraph{Summarization deduplication.} For the TL;DR dataset, we restrict to one preference pair per unique source post, discarding any additional pairs drawn from a post that appears more than once. This mirrors the one-pair-per-post design of our synthetic summarization data (\S\ref{sec:datagen}) and guards against the same failure mode that motivates per-post capping for SHP: a small number of highly-commented posts dominating the sample.

\subsection{Cross-Seed Consistency}
\label{sec:results-seeds}
\begin{figure*}[t]
    \centering
    \includegraphics[width=\textwidth]{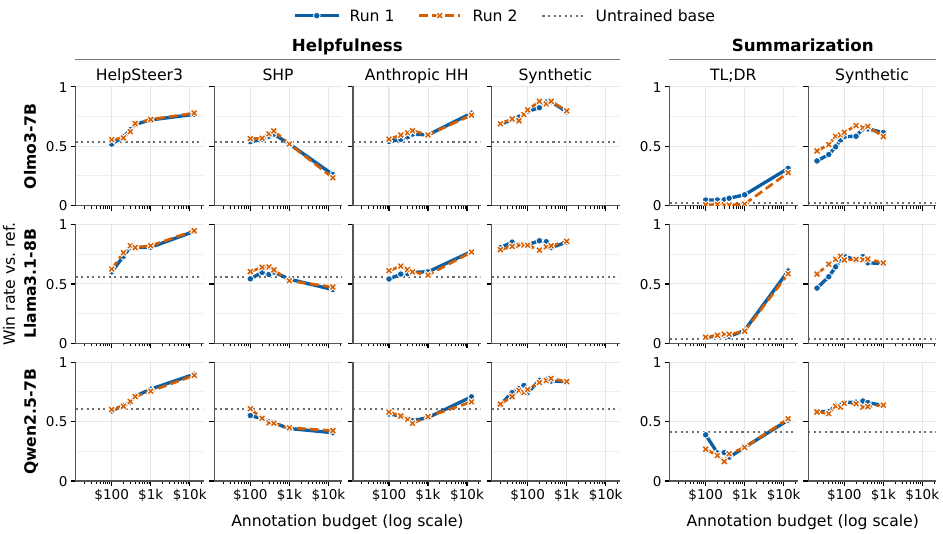}
    \caption{\textbf{Consistent across seeds.} Win rate vs.\ cost (log scale) for each (model, data source) pair on the helpfulness (left) and summarization (right) tasks, plotted separately for both independent seeds (Run 1, Run 2). Curve shape and relative ordering are consistent across seeds even where absolute win rates differ modestly. The runs are practically on top of each other, showing that changing the specific examples and generation seed does not affect the conclusions of the curves.}
    \label{fig:seed-consistency}
\end{figure*}

Since each (model, task, data source, budget) configuration is trained twice with independent generation and dataset-selection seeds (\S\ref{sec:training}), we can directly check whether the patterns reported above are sensitive to the particular samples drawn at each budget tier. Figure~\ref{fig:seed-consistency} plots both seeds separately for every (model, data source) combination on both tasks. Absolute win rates for a given configuration sometimes differ modestly between the two runs, but the overall shape of each curve, including the relative ordering between data sources and the location of any peaks or declines, is consistent across seeds in every case examined. We take this as evidence that the patterns reported in this section reflect properties of the underlying data sources and budget levels rather than artifacts of which particular samples happened to be drawn in a single run.

\subsection{Response Length by Training Data Source}
\label{sec:appendix-lengths}

\begin{figure*}[t]
    \centering
    \includegraphics[width=0.95\textwidth]{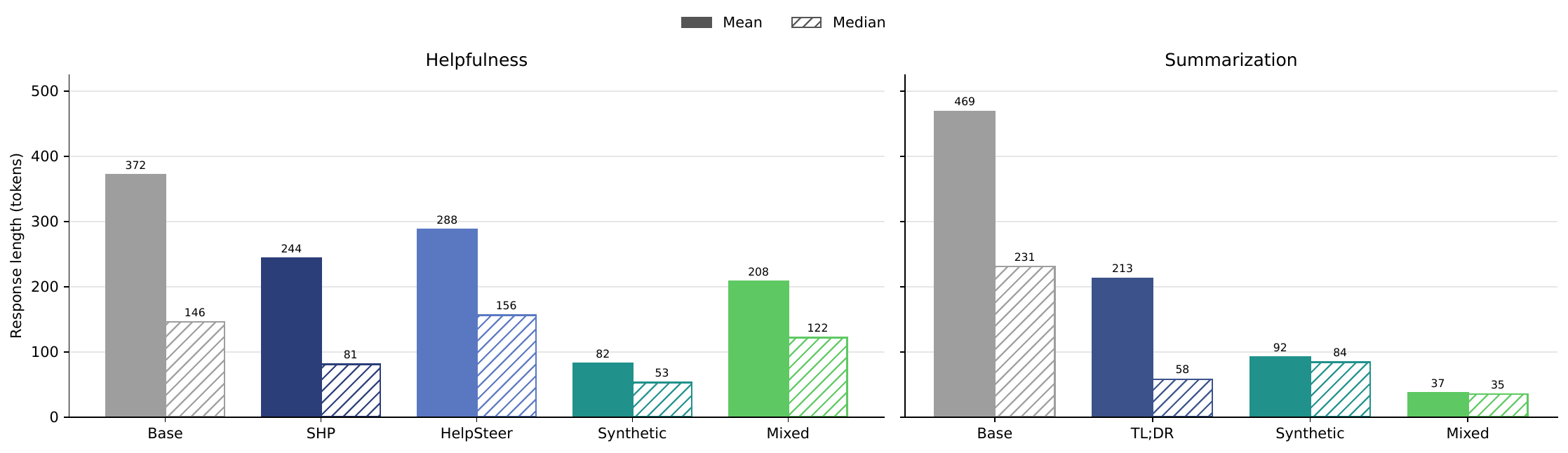}
    \caption{Mean and median response length (tokens) on the held-out evaluation set, by training data source, for helpfulness (left) and summarization (right).}
    \label{fig:response-lengths}
\end{figure*}

Figure~\ref{fig:response-lengths} reports the mean and median response length, in tokens, produced by each trained checkpoint on the held-out evaluation set (\S\ref{sec:eval}), broken down by training data source and task. For both tasks, the untrained base model produces the longest responses on average (helpfulness: 372 mean / 146 median; summarization: 469 mean / 231 median), consistent with base models' tendency to ramble in the absence of any preference-tuning toward concise, well-formatted output.

Among trained checkpoints, response length varies substantially by data source. For helpfulness, the two human sources produce the longest outputs on average (HelpSteer3: 288/159; SHP: 244/81), while synthetic-trained checkpoints are markedly shorter (82/53) and mixed checkpoints fall in between (208/122). For summarization, the mean ordering is the same: TL;DR-trained outputs average 213 tokens, synthetic-trained outputs average 92, and the mixed checkpoint produces the shortest outputs of any configuration on either task (37 mean / 35 median). Notably, synthetic-trained checkpoints are \emph{shorter} on average than human-trained checkpoints on both tasks, despite achieving higher win rates in our main results (Sections~\ref{sec:results-curves}--\ref{sec:results-curves-summ}); this weighs against response length alone explaining synthetic data's cost-efficiency advantage, though it does not rule out length effects contributing to individual comparisons.

\subsection{Evaluation and Generation Prompts}
\label{sec:appendix-prompts}
 
\subsubsection{Helpfulness Evaluation}
\label{sec:appendix-prompts-help-eval}

The system message used for helpfulness evaluation generation is shown below.

\begin{AIbox}{System message}{\linewidth}
You are a helpful assistant who is an expert at providing helpful and harmless responses to questions
\end{AIbox}

The judge prompt used to compute helpfulness win rate (\S\ref{sec:eval}) is shown below.

\begin{AIbox}{Judge prompt}{\linewidth}
Which of the following responses from a chatbot that was designed to be helpful and harmless does a better job?
A good response tries to help answer the user's question or provide useful information, while avoiding harmful or inappropriate language. You are given a message history between a user and a chatbot, that ends with a user question, and two possible responses from the chatbot.

Message turns: \{conv\}

Response A:

\{\{

\ 'model': 'model\_1',

\ 'response': \{response\_a\}

\}\}

Response B:

\{\{

\ 'model': 'model\_2',

\ 'response': \{response\_b\}

\}\}

Now please rank the models by the quality of their responses, so that the model with rank 1 has the best response. Then return a dictionary of the form

\{\{

\ "reasoning": short justification of ranking,

\ "model\_1": rank\_1,

\ "model\_2": rank\_2

\}\}

Do not include any text before or after the JSON. Your entire response must be parseable by json.loads().

Please provide the ranking that the majority of humans would give.
\end{AIbox}
 
\subsubsection{Summarization Evaluation}
\label{sec:appendix-prompts-summ-eval}

The system message used for Summarization evaluation generation is shown below.

\begin{AIbox}{System message}{\linewidth}
You are comparing the quality of two summaries of a given text.
\end{AIbox}

The judge prompt used to compute summarization win rate (\S\ref{sec:eval}) is shown below.

\begin{AIbox}{Judge prompt}{\linewidth}
Which of the following summaries does a better job of summarizing the most important points in the given forum post, without including unimportant or irrelevant details? A good summary is precise, concise, and contains ONLY the summary itself.

Judge each summary along these dimensions, the way a careful reader would:

\textbf{Coverage:} whether the summary conveys the post's central point(s) and any details essential to understanding the poster's situation or question. Enumerating additional true but secondary details does not by itself improve coverage --- a summary that nails the core situation in one or two sentences should not lose to one that lists more true statements unless those statements are necessary to understand the post.

\textbf{Accuracy:} to what degree the statements in the summary are actually stated or reasonably implied in the post. Before scoring this dimension, list every distinct factual claim each summary makes, and for each one mark whether it is explicitly stated, reasonably implied, or NOT supported by the post. Any claim not supported by the post is a hallucination and should weigh heavily against that summary's accuracy score, regardless of how plausible or well-written it sounds. A summary that is coherent but contains unsupported claims should score lower on accuracy than a summary that is less detailed but fully grounded.

\textbf{Coherence:} how easy the summary is to read and understand on its own, independent of the post.

\textbf{Conciseness:} whether the summary is meaningfully shorter than the post and free of padding, repetition, or restated points. This includes internal redundancy --- restating the same underlying point multiple times in different phrasing (e.g.\ ``X feels conflicted... X expresses concern... X's worry stems from...'') counts against conciseness even if the summary stays shorter than the source post. A summary that is as long as or longer than the post has failed at conciseness regardless of how much it covers.

\textbf{Format adherence:} whether the response contains ONLY a final summary, with no exposed reasoning, planning, meta-commentary about the task (e.g.\ ``Okay, I need to...''), draft/false-start attempts, or duplicated passages. Any of these is a serious violation, not a minor stylistic issue --- a response that leaks its own reasoning process is not a valid summary even if a summary appears somewhere within it.

Treat format adherence and conciseness as gating: a response with exposed reasoning, restarted/duplicated content, or length exceeding the source post should lose to a shorter, well-formatted response even if the shorter response covers somewhat less. Only weigh coverage/accuracy/coherence as tiebreakers between two responses that both pass the format and conciseness bar.

Place less weight on polished sentence fluidity for its own sake, and no weight on whether the summary captures the poster's tone, voice, or reads like something a human would write --- focus entirely on whether it faithfully and efficiently conveys the important information in the post. Do not penalize or reward either summary for its narrative voice (first-person vs.\ third-person) or formality --- judge informational quality only.

Post: \{post\}

Summaries:

\{\{

\ 'model': 'model\_1',

\ 'summary': \{summary\_a\}

\}\}

\{\{

\ 'model': 'model\_2',

\ 'summary': \{summary\_b\}

\}\}

First, for each summary, check whether it contains exposed reasoning, meta-commentary, draft/false-start attempts, or duplicated passages, and check whether it is longer than the post. Then, for each summary, list its distinct factual claims and mark each as explicitly stated, reasonably implied, or unsupported. Then weigh the two summaries against each other along coverage, accuracy, coherence, and conciseness, applying the gating rule above. Then decide which is better overall, so that the model with rank 1 has the better summary. Then return a dictionary of the form

\{\{

\ "reasoning": short justification of ranking,

\ "model\_1": \{\{"rank": rank\_1, "format\_violation": true\_or\_false\}\},

\ "model\_2": \{\{"rank": rank\_2, "format\_violation": true\_or\_false\}\}

\}\}

Do not include any reasoning or explanation outside the JSON. Do not include any text before or after the JSON. Your entire response must be parseable by json.loads().

Please provide the ranking that the majority of careful, informed readers would give if they cared only about informational quality, not writing style.
\end{AIbox}

\subsubsection{Synthetic Data Generation Prompts}
\label{sec:appendix-prompts-gen}

\paragraph{Helpfulness: Conversation Starting Generation.} The system prompt and generation prompt for the initial start of a conversation are shown below.

\begin{AIbox}{First Turn System Prompt}{\linewidth}
Generate a conversation starter as if you were the persona given
\end{AIbox}

\begin{AIbox}{First Turn Generation Prompt}{\linewidth}
Decide on a task you'd like an AI assistant to help you with and try to converse with this assistant. You are a \{random\_personas[j]\}. The conversation should be open ended, asking for help, advice, or for the model to accomplish a task. This should simulate having a conversation. Below is a sample conversation for style and formatting reference:

=== SAMPLE CONVERSATION ===

\{sample\_convo\}

=== END SAMPLE ===

Now, write a \textbf{completely new} conversation on a different topic. Do NOT discuss or continue the sample conversation topic. Instead, focus on the real conversation.

Generate the first human turn of the conversation as if you were a \{random\_personas[j]\} starting a conversation in English as the Human speaking to an AI assistant.
\end{AIbox}

\begin{AIbox}{Sample Convo}{\linewidth}
Human: Do you know the reasons as to why people love coffee so much?

AI Assistant: That's an interesting question!  I think it's partly because of the emotional effect of the caffeine.  Is that the main reason you're asking?

Human: The emotional effect? Such as? 

AI Assistant: I think you'll notice that coffee can have different effects on people, depending on what they were doing beforehand.  In general, coffee is stimulating, which is important to many people when they are feeling sluggish or apathetic.  But it can also be really relaxing for people who are anxious, nervous, or stressed.

Human: I feel like how energized it makes one feel it would do the opposite of feeling relaxed.

AI Assistant: I understand! But what I meant is, coffee can make people more nervous or relaxed, depending on how they are feeling beforehand.  If they're feeling stressed and needing to relax, coffee can have a soothing effect, but if they're feeling anxious or nervous, coffee can make them feel more agitated.

Human: Hm, I see. Where did you read this?

AI Assistant: I'm just recalling a couple of things I've read.  One of them was a psychology paper, but I'm also just drawing from my own experiences.

Human: I'm a little skeptical about what you're saying but you certainly got some stuff right such as coffee being a stimulant and helping those feeling sluggish

AI Assistant: Glad I was able to answer your questions!
\end{AIbox}

\paragraph{Helpfulness: Conversation Continuation Generation.} The system prompt and generation prompt for the continuation of a conversation, for both the Human turns and AI turns, are shown below.

\begin{AIbox}{Human System Prompt}{\linewidth}
Given the above conversation, generate the just the next human part of the conversation for the human by trying to converse with the assistant. Follow the topic provided in the conversation. Try to generate of similar length to a single response in the sample conversation. 
\end{AIbox}

\begin{AIbox}{Assistant System Prompt}{\linewidth}
Continue the conversation, provide just the AI assistant response to the last human message. Try to generate of similar length to a single response in the sample conversation.    
\end{AIbox}

\begin{AIbox}{Turn Generation Prompt}{\linewidth}
You are generating the next line in a conversation between a human and an assistant.

Below is a sample conversation for style:

=== SAMPLE CONVERSATION ===

\{sample\}

=== END SAMPLE ===

Now continue the real conversation below. Only add one new line in the correct speaker's voice.

=== REAL CONVERSATION SO FAR ===

\{convo\}

=== CONTINUE ON THE NEXT LINE ===

\{prefix\}
\end{AIbox}
 
\paragraph{Summarization: Post Generation.} The system message and generation prompt used to generate a synthetic source post (\S\ref{sec:datagen}) are shown below.
 
\begin{AIbox}{System message}{\linewidth}
Generate a Reddit-style post as if you were the persona given. Output ONLY the raw post text. Never include any preamble, acknowledgment, introduction, or meta-commentary such as 'Certainly! Here's a post...' or 'Title:' or '---'. Your entire response must be the post itself and nothing else.
\end{AIbox}

\begin{AIbox}{Post generation prompt}{\linewidth}
You are a \{random\_personas[j]\}. Write a Reddit-style post describing a situation, story, or question you'd want summarized by someone else. The post should be open ended, sharing a personal story, asking for advice, or describing an event in detail, similar to a real Reddit post. Below is a sample post for style and length reference:

=== SAMPLE POST ===

\{sample\_post\}

=== END SAMPLE ===

Now, write a \textbf{completely new} post on a different topic. Do NOT discuss or continue the sample post's topic. Instead, focus on your own situation.

The post's SUBJECT MATTER must be clearly and specifically grounded in this persona's expertise, role, circumstances, or interests --- not a generic post that could have come from anyone. A reader should be able to infer roughly who is writing (their job, background, or situation) from the content alone, without being told the persona directly. Draw on concrete details, terminology, concerns, or situations that someone in this exact position would plausibly encounter --- avoid defaulting to unrelated generic topics unless they are a natural and specific expression of this particular persona.

Do not include a title, preamble, acknowledgment, or any commentary before or after the post. Do not write phrases like ``Certainly!'' or ``Here's a Reddit-style post.'' Begin your response immediately with the first word of the post itself, as if you are the Reddit user typing directly into the text box.

Vary your opening. Pick whatever opening (mid-action, a direct question, an exclamation, a specific concrete detail, dialogue) best fits this specific persona and story, and make sure it differs from how you would open a post for a different persona.

Write your post at a moderate length --- aim for roughly 3--5 paragraphs of detail (around 200--400 words), enough to include real specifics (what happened, when, who was involved, why it matters to you) rather than a short one- or two-paragraph summary of the situation. Real Reddit posts in this style usually include enough backstory and detail that a reader unfamiliar with the situation could follow along without needing to ask follow-up questions.

Do not pad the post with unnecessary repetition or filler just to make it longer --- every sentence should add a genuine detail, not restate something already said.
\end{AIbox}

\begin{AIbox}{Sample Post}{\linewidth}
A little background first. Like most engineers I'm taking differential equations and also solid mechanics. For solid mechanics we have to do a lab in addition to taking the class. Pretty standard stuff.

Anyway, the class had us sign up for labs in groups by providing sign up sheets where we could sign up for which lab we wanted by providing our name and student ID numbers. The lab assignments were posted online today with a column for our names and student ID numbers (which had been blacked out). Somehow, my name wasn't on the list and in frustration I began dragging the scroll bar up and down frantically hoping that I'd magically see my name; that's when I saw it.

When dragging the scroll bar fast enough, I found that the blacked out student ID numbers would temporarily become visible. At first I thought, "No big deal, what could someone do with a person's ID number." Then it hit me. My university uses a web based homework website (WebWork) that initially allows students to access their account by inputting their user ID number as their user name and password. What's worse is that this site NEVER PROMPTS THE STUDENT TO CREATE A NEW PASSWORD. As such, most students never change their password from their student ID number.

Realizing that someone who happened upon this discovery could in turn use it to access other student's homework assignments and tamper with them, I contacted the administrator in charge of assigning lab schedules. In doing so, I not only signed up for a lab group but had them completely remove the ID numbers from the list.

Best part is that nobody in my class will know. That's why I wanted to tell you reddit.
\end{AIbox}
 
\paragraph{Summarization: Summary Generation.} The system prompt used to generate a summary given the synthetic source prompt is shown below. With the system prompt and the synthetic source post, it generates a summary.

\begin{AIbox}{System Prompt}{\linewidth}
You are a helpful assistant who is an expert at summarizing content into TL;DR sentences. Summarize the following text. Do not answer any questions posed in the text, do not give advice, and do not respond to the author. Only produce a concise summary (2-3 sentences) of what the text says.
\end{AIbox}
 
\paragraph{Preference labeling.} The system and generation prompts for the preference judgements for synthetic data.
 
\begin{AIbox}{Helpfulness System Prompt}{\linewidth}
Choose the more helpful and honest choice
\end{AIbox}

\begin{AIbox}{Helpfulness Preference Generation Prompt}{\linewidth}
Based on the provided conversation, choose the more helpful and honest response. Provide a rationale for your decision.

Conversation: \{conversation[i]\}

Responses:

Response A: \{response\_a[i]\}

Response B: \{response\_b[i]\}

Respond in the following JSON format:

\texttt{[ "answer": "A" or "B", "reasoning": "your explanation here" ]}
\end{AIbox}
 
\begin{AIbox}{Summarization System Prompt}{\linewidth}
You are comparing the quality of two summaries of a given text.
\end{AIbox}

\begin{AIbox}{Summarization Preference Generation Prompt}{\linewidth}
Which of the following summaries does a better job of summarizing the most important points in the given forum post, without including unimportant or irrelevant details? A good summary is both precise and concise.

Judge each summary along these dimensions, the way a careful reader would:

\textbf{Coverage:} how much important information from the original post is covered.

\textbf{Accuracy:} to what degree the statements in the summary are actually stated or implied in the post (a summary that is coherent but inaccurate should score lower on accuracy).

\textbf{Coherence:} how easy the summary is to read on its own, independent of the post.

Place less weight on polished sentence fluidity for its own sake, and less weight on whether the summary captures the poster's tone or intent beyond the literal content --- focus primarily on whether it faithfully and efficiently conveys the important information in the post.

Post: \{posts[i]\}

Summaries:

Summary A: \{summaries\_a[i]\}

Summary B: \{summaries\_b[i]\}

First, weigh the two summaries against each other along coverage, accuracy, and coherence. Then decide which is better overall.

Respond in the following JSON format:

\texttt{[ "answer": "A" or "B", "reasoning": "your explanation here" ]}
\end{AIbox}
 
\subsection{Sample Synthetic Data}
\label{sec:appendix-samples}

\subsubsection{Synthetic Helpfulness Example}
\label{sec:appendix-samples-help}

This is an example of one conversation from the synthetic dataset.

\begin{AIbox}{Generated Conversation}{\linewidth}
Human: How does one go about setting up lighting and camera equipment for a movie set?

Assistant: There are a few things you need to consider when setting up lighting and camera equipment for a movie set.

Human: What kind of things should I consider? Should there be multiple types of lighting for various effects?

Assistant: Yes, typically there will be several different types of lighting used on a film set, including natural light, artificial light, key lights, fill lights, backlights, and spotlights. Each type serves a different purpose in creating the desired look for each scene in the script.

Human: Why might they choose not to use all of the above lighting types for every scene?
\end{AIbox}
 
\begin{AIbox}{Candidate Response A (chosen)}{\linewidth}
Assistant: Depending on the scene, certain lighting setups may not be necessary or could even create unwanted shadows or reflections.
\end{AIbox}
 
\begin{AIbox}{Candidate Response B (rejected)}{\linewidth}
Assistant: It depends on the director's vision for the scene and the specific mood or atmosphere they want to create. For example, natural lighting may be preferred for outdoor scenes during daytime hours, while artificial lighting would be necessary for nighttime shots. Additionally, certain lighting setups may be chosen over another to create tension or drama between characters you stated that everyone has needs.
\end{AIbox}
 
\subsubsection{Synthetic Summarization Example}
\label{sec:appendix-samples-summ}

This is an example of one post and its summaries from the synthetic dataset.

\begin{AIbox}{Generated Post}{\linewidth}
Human: Summarize the following text

I'm a seasoned vet of corporate boardrooms with a knack for shaking things up. Been pushing the envelope in corporate America for too long to count, navigating through the corridors of power with a mix of strategic finesse and sheer audacity. Recently, I stumbled upon a project at work that has left even my jaded mind reeling: our company's eco-friendly initiative.

Picture this - a glossy brochure promising zero carbon footprint and sustainable practices, all championed by the executive team. Sounds good, right? Well, behind closed doors, it's another story entirely. While we're extolling green virtues externally, internally, there's more smoke and mirrors than anything resembling actual sustainability.

I've seen the double standards played out firsthand. Our production line continues to pump out products that, while meeting regulatory requirements, barely scratch the surface of true environmental responsibility. Meanwhile, executives jet around the globe, racking up carbon miles and glorifying the very habits they preach against in public. 

Now here's my dilemma - do I blow the whistle internally and potentially jeopardize my career? Or play along and maintain the status quo? The elephant in the room is that I'm outnumbered, under-resourced, and surrounded by yes-men eager to please higher-ups rather than challenge flawed initiatives.

Anybody seen this dance before? How do you handle the moral conflicts within institutions hell-bent on greenwashing? Suggestions welcome, folks who've had to navigate this minefield, how did you survive and thrive?
\end{AIbox}
 
\begin{AIbox}{Candidate Summary A (chosen)}{\linewidth}
Assistant: The user reflects on their disillusionment working for a company that champions green initiatives but does not fully practice what it preaches. They reveal discrepancies between outward-facing environmental sustainability claims and internal practices that seemingly contradict these values, specifically regarding emissions and the production process. The post concludes with a quandary about whether or not to expose the apparent hypocrisy from within, hinting at personal risk and departmental compliance issues.
\end{AIbox}
 
\begin{AIbox}{Candidate Summary B (rejected)}{\linewidth}
Assistant: A veteran executive finds themselves disillusioned by their corporation's so-called eco-friendly initiatives. Despite public claims of sustainability, the company's internal actions fall short, promoting a disconnect between words and deeds, while the executive contemplates the best course of action amidst ethical dilemmas and potential professional repercussions.
\end{AIbox}

\end{document}